%% file: acl_latex.tex
\pdfoutput=1

\documentclass[11pt]{article}

\usepackage[final]{acl}

\usepackage{times}
\usepackage{latexsym}

\usepackage[T1]{fontenc}

\usepackage[utf8]{inputenc}

\usepackage{microtype}

\usepackage{inconsolata}

\usepackage{graphicx}

\usepackage[utf8]{inputenc} 
\usepackage[T1]{fontenc}    
\usepackage{hyperref}       
\usepackage{url}            
\usepackage{booktabs}       
\usepackage{amsfonts}       
\usepackage{nicefrac}       
\usepackage{microtype}      
\usepackage{xcolor}         
\usepackage{multirow,multicol}
\usepackage{graphicx}

\usepackage{enumitem}

\title{Incubating Text Classifiers Following User Instructions\\with Nothing but LLM}

\author{
  Letian Peng, Jingbo Shang\thanks{$\ $  Corresponding author.}\\
  Department of Computer Science\\
  University of California, San Diego\\
  \texttt{\{lepeng, jshang\}@ucsd.edu} \\
}

\usepackage{xspace}

\newcommand{\our}{Incubator\xspace}

\begin{document}
\maketitle
\begin{abstract}
    \input{0-abs}
\end{abstract}

\input{1-intro}

\input{2-rel}
\input{3-method}
\input{4-exp}

\input{5-analysis}

\input{6-con}
\input{7-lim}

\bibliography{custom}

\clearpage

\appendix

\section{Hyperparameter}
\label{apdx:hyperparameter}

\begin{table}[h]
\small
\centering
\scalebox{1.0}{\begin{tabular}{lcc}
\toprule
Hyperparameter & Instruction-tuning & Incubation \\
\midrule
Initial LR & $2\times 10^{-5}$ & $1\times 10^{-5}$ \\
Batch Size & $16$ & $32$ \\
Epoch & $3$ & $8$ \\
\bottomrule
\end{tabular}}
\caption{The hyperparameter setups in our experiments.} 
\label{tab:hyperparameter}
\end{table}

\section{Instruction-tuning Dataset Processing}
\label{apdx:convert}

\begin{figure}[h]
    \centering
    \includegraphics[width=\linewidth]{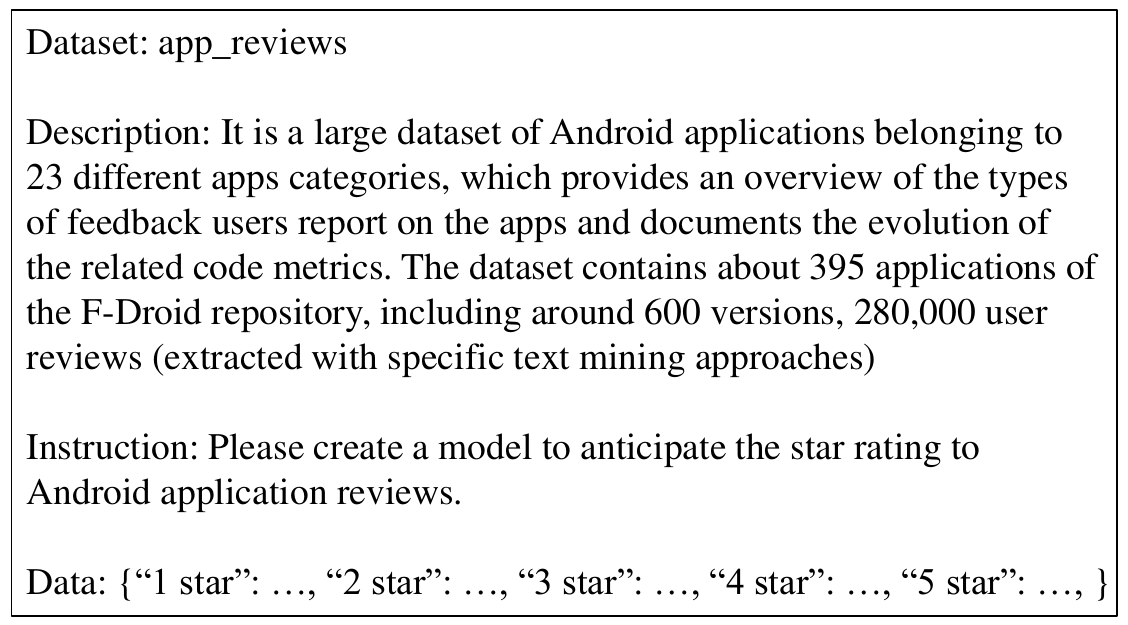}
    \caption{A case in our instruction-tuning dataset for \our.}
    \label{fig:process}
\end{figure}

\section{Dataset Generation Prompt}
\label{apdx:icla}

\begin{table}[h]
\centering
\small
\scalebox{1.0}{
\begin{tabular}{p{1.5cm}p{5cm}}
\toprule
\textbf{Role} & \textbf{Message} \\
\midrule
User & Generate an imaginative instruction to build a text classifier and its corresponding samples. \\
\midrule
GPT-4 & ``Input'': ``Instruction 1'' \\ 
& ``Output'': \{``Label 1,1'': ``Data 1,1'',\\
& ``Label 1,2'': ``Data 1,2'', ...\}\\
\midrule
User & Generate an imaginative instruction to build a text classifier and its corresponding samples. \\
\midrule
GPT-4 & ``Input'': ``Instruction 2'' \\ 
& ``Output'': \{``Label 2,1'': ``Data 2,1'',\\
& ``Label 2,2'': ``Data 2,2'', ...\}\\
\midrule
User & Generate an imaginative instruction to build a text classifier and its corresponding samples. \\
\bottomrule
\end{tabular}
}
\caption{The prompt used in ICL-based augmentation.} 
\label{tab:icla}
\end{table}

\newpage

\section{Revised Dataset with Miscellaneous} 
\label{apdx:misc}

\begin{table}[h]
\small
\centering
\scalebox{0.8}{\begin{tabular}{lll}
\toprule
Dataset & Label & Other \\
\midrule
Emotion & Joy, Sadness & Love, Anger, Fear, Surprise \\
\midrule
\multirow{2}*{NYT-LOC} & America, Iraq, & Britain, German, Canada, \\
 & Japan, China & France, Russia, Italy \\
\midrule
\multirow{3}*{Massive} & Calendar, Play, & Lists, News, Recommendation, \\
 & QA, Email, IoT, & Datetime, Social, Alarm, Music, \\
 & Weather, Transport & Audio, Takeaway, Cooking \\
\bottomrule
\end{tabular}}
\caption{The revision on datasets for the label ``Other''.} 
\label{tab:misc_data}
\end{table}

As shown in Table~\ref{tab:misc_data}, the minor categories with low proportion are merged together to an ``Other'' class.

\end{document}

%% file: 0-abs.tex


In this paper, we aim to generate text classification data given arbitrary class definitions (i.e., user instruction), so one can train a small text classifier without any human annotation or raw corpus.
Compared with pioneer attempts, our proposed \our is the first framework that can handle complicated and even mutually dependent classes (e.g., ``\emph{TED Talk given by Educator}'' and ``\emph{Other}'') . 
Specifically, \our is an LLM firstly tuned on the instruction-to-data mappings that we obtained from classification datasets and descriptions on HuggingFace together with in-context augmentation by GPT-4.
We then
refine \our by learning on the cluster centers of semantic textual embeddings to emphasize the uniformity and semantic diversity in generations.
We compare \our on various classification tasks with strong baselines such as direct LLM-based inference and training data generation by prompt engineering.
Experiments show \our is able to (1) perform well on traditional benchmarks, (2) take label dependency and user preference into consideration, and (3) enable logical text mining by incubating multiple classifiers.
\footnote{Code: \href{https://github.com/KomeijiForce/Incubator}{KomeijiForce/Incubator}}

%% file: 1-intro.tex
\section{Introduction}

Text classification plays a 
vital 
role in many natural language processing (NLP) systems, 
such as email system~\citep{spam-system}, text mining~\citep{text-mine-system}, and recommender systems~\citep{book-recommendation-system}. 

Different from the traditional supervised way to build a text classifier, which fine-tunes models on human annotations~\citep{ag_news},
zero-shot text classification reduces manual effort by building classifiers with minimal inputs, such as label names~\citep{x-class, sci-class, wot-class}.
These zero-shot methods are typically based on mining pseudo training data from massive raw texts with precise filtering algorithms,
which unfortunately limits their application to simple labels. 
For more complex labels, their distributions are extremely scarce in raw texts and filtering algorithms struggle to recall these examples while maintaining their precision. 

Large language models (LLMs), such as GPT-3~\citep{gpt-3}, have been recently introduced to address this problem with their proficient capability to capture the nuance in complex labels. 
Specifically, people prompt LLMs to generate data based on 
each label,
and then fine-tune small classifiers as the final production~\citep{zerogen,progen}.


\begin{figure}[t]
    \centering
    \includegraphics[width=\linewidth]{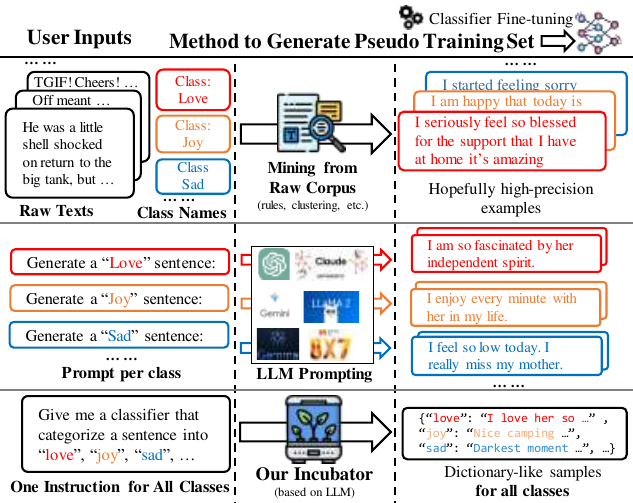}
    \vspace{-3mm}
    \caption{A comparison of \our with different methods for zero-shot text classification.}
    \label{fig:intro}
    \vspace{-5mm}
\end{figure}

\begin{figure*}[t]
    \centering
    \includegraphics[width=\linewidth]{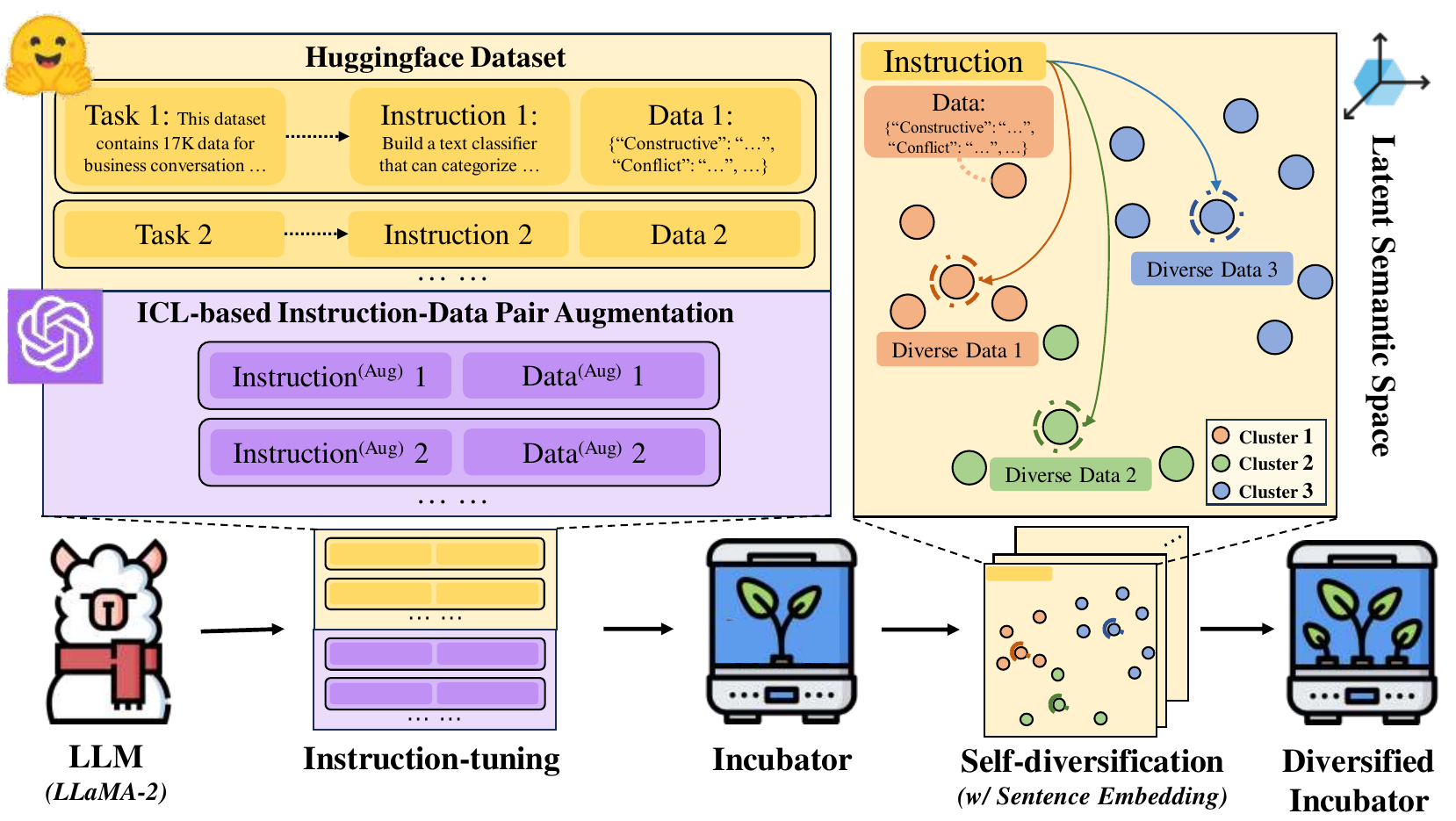}
    \vspace{-3mm}
    \caption{An overview of our framework to build \our.}
    \label{fig:overview}
    \vspace{-5mm}
\end{figure*}

Existing LLM-based zero-shot text classification methods, while feasible, face two major challenges,
(1) class definitions can go beyond a simple label name, such as ``\emph{TED Talk given by Educator}''
and (2) class definitions can depend on each other.
For example, the class ``\emph{Other}'' is only meaningful when seeing other classes; 
As shown in Figure~\ref{fig:intro}, the class ``\emph{Optimistic}'' shall not contain ``\emph{Love}'' when ``\emph{Love}'' itself presents as a class.
Therefore, the scope of the class with the same textual definition can vary as other classes change.

We argue that the LLMs need further instruction-tuning~\cite{instructgpt}, particularly for classification data generation.
Specifically, we leverage public classification datasets with descriptions for this tuning.
This allows the user to control the LLM to generate useful training data for small models based on (1) label interdependency and (2) user preferences described in the instructions.
Consequently, the LLM-based zero-shot text classification can be simplified as \emph{model incubation} that ``User requires a model by an instruction, the LLM (\textbf{\our}) then generates useful training data to incubate such a model.''

In this paper, we first collect some pairs of dataset descriptions and training data samples on Huggingface~\citep{huggingface}, each formalized as a dictionary with each label as a key and a sample as the value. 
These data are beneficial for \our to learn label interdependency 
as the examples from different classes are presented jointly.
Then the data descriptions are manually converted to user instructions, which establishes a mapping from the user instruction to training data. 
These instructions are augmented by a very strong LLM (e.g., \texttt{GPT-4})
using in-context learning (ICL) \cite{icl-survey} and used to instruction-tune an open-source LLM (e.g., \texttt{LLaMA-2-7b-hf})
as our \our. 
Note that we can leverage \texttt{GPT-4} with ICL as \our too.
We recommend open-source LLMs as \our because of open parameters, inference efficiency, and further fine-tuning.


To alleviate the known negative impact of data bias on text classification~\cite{text-bias,BFClass,WeDef} and bias in contents generated by LLMs~\citep{llm-bias,bias-ai-content}, we propose a novel self-diversification technique to increase the data uniformity and diversity, utilizing the text representations from a text embedder~\cite{E5}. 
Specifically, we instruct the \our many times (e.g., $1024$), and then use a clustering algorithm (e.g., K-means) to get the sample nearest to each cluster center, which is considered very semantically different from one another. 
These samples are incorporated in the same batch to further instruct-tune \our to increase the data uniformity and diversity.

We conduct experiments to test the instruction-following ability of our \our on various tasks to test its basic incubation ability, label interdependency awareness, and user instruction following ability. 
These tasks involve incubating text classifiers for (1) traditional benchmarks, (2) classification tasks with ``\emph{Other}'' as a label, and (3) classification tasks with user customization for personal preference. 
We include strong baselines such as directly instructing the LLM to classify texts and prompting LLMs to generate data for each label separately. 

Experiment results verify our \our to be able to (1) incubate strong text classifiers that outperform the baselines, (2) consider the label interdependency and follow the user preference in the instruction, (3) incubate multiple text classifiers and use logical conjunctions to realize advanced text mining systems.

Our contributions in this paper are three-fold.
\begin{itemize}[nosep,leftmargin=*]
    \item We propose the first instruction-tuning framework to learn the LLM as an \our, which incubates text classifiers following user instructions.
    \item We propose a novel self-diversification technique, which utilizes the cluster centers of generated samples to increase the uniformity and diversity in \our generation.
    \item We conduct extensive experiments on benchmark datasets to demonstrate the superior accuracy of the incubated text classifiers.
    \item We showcase how to apply \our to realize advanced text mining systems by incubating multiple text classifiers with logical conjunctions\footnote{The datasets and models used in the experiments will be released for reproductivity.}.
\end{itemize}

%% file: 2-rel.tex
\section{Related Works}

\subsection{Zero-shot Text Classification}

Traditional zero-shot text classification methods are based on text mining in massive raw texts with label names \citep{x-class, sci-class, wot-class}. A related setup allows incorporating some seed words for each class to strengthen the text mining precision \citep{xws-benchmark,debiase-sota}. With the emergence of LLMs, many pioneer studies on LLM-based zero-shot text classification propose to prompt LLMs with label names and fine-tune small classifiers on those generated results \citep{zerogen, progen}. However, these methods are substantially label-wise text generation, which fails to consider the whole classification task, involving label interdependency and user preference. Our work aims to fill in this blank by proposing a framework that builds customized classifiers according to user instructions.

\subsection{Instruction-tuning}

Following instructions~\cite{instruction-tuning-survey} is a fundamental capability for Large Language Models (LLMs), crucial for understanding and acting upon user commands, thus enhancing their appeal to user-specific applications. InstructGPT~\cite{instructgpt} represents an initial exploration into LLMs tailored to follow instructions, revealing their capacity to perform tasks as directed by users. ChatGPT~\cite{gpt-4}, with its superior capability to follow instructions, bolstered by reinforcement learning with human feedback (RLHF), has enjoyed considerable acclaim both within and beyond the language research community. Furthermore, publicly available LLMs designed for instruction-following, such as LLaMA~\cite{llama,llama2}, offer a rich foundation for investigating the ability of LLMs to execute instructions. Instruction-tuning not only contributes to the success of LLMs in text-to-text tasks \citep{instruction-tuning-survey}, but is also able to customize image generation \cite{instruction-text-to-image} and text embeddings \cite{inbedder}. Our work follows this trend to instruction-tune LLMs as \our, which customize classifiers according to user instructions. 

\subsection{Model Incubation}

The area closest to model incubation is symbolic distillation \cite{symbolic-distill,symbolic-cot-distill}, which distills a teacher model into a different type of student model. Those student models can function very differently from the initial language modeling teacher, such as commonsense reasoning \cite{symbolic-distill} and information extraction \cite{universalner}. Another relevant domain is training data generation including augmentation. Besides classification data generation \citep{zerogen,progen,cotam}, there also exists generation pipelines for question answering \cite{QG-Conv,QG-RAG} and natural language generation \cite{nlg-aug}. Model incubation differs from previous works as it takes user instruction as the input, which allows a more user-oriented model customization for personal usage. 

%% file: 3-method.tex
\section{Our \our Framework}


Figure~\ref{fig:overview} offers an overview of our \our framework, including two stages, (1) \textbf{Instruction-tuning} and (2) \textbf{Self-diversification}. 
The instruction-tuning stage utilizes the existing resources on the Huggingface platform to learn an LLM as \our to generate training data based on user instructions. 
The self-diversification stage further improves the uniformity and diversity in \our generation with an auxiliary text embedder and clustering. 
We now elaborate on the details of these two stages.

\subsection{Instruction-tuning for \our}

\paragraph{Instruction-to-data Dataset} 
We select $25$ text classification datasets on the Huggingface platform\footnote{\href{https://huggingface.co/datasets}{https://huggingface.co/datasets}} to build the initial instruction-to-data dataset for instruction-tuning, such as financial news, counterfactual reviews, and toxic conversations. 
For each dataset, we extract its description and sample a few (we select $10$) samples per class from it, which are formalized as Python dictionaries.
The keys in the dictionary are labels and each label corresponds to one text data as the value. 
Consequently, we get $250$ instruction-to-data samples as the initial dataset. 
We present a specific case inside the dataset in the Appendix~\ref{apdx:convert}.

\paragraph{ICL-based Augmentation} 
Directly instruction-tuning the LLM on the initial dataset will likely introduce overfitting and bias to the \our due to the limited number of instructions \citep{few-shot-survey}. 
Thus, we address these issues by data augmentation \cite{spurious-correlation-survey} and use ICL \cite{icl-survey} by \texttt{GPT-4} \cite{chatgpt} as the implementation \cite{icl-distillation}. 
We show the specific prompt for in-context learning in Table~\ref{tab:icla} of Appendix~\ref{apdx:icla}. 
We have two in-context examples that map instructions to training data as Python dictionaries, which are randomly sampled in each query. 
Finally, we augment the instruction-to-data dataset to $12$K samples.
This dataset is then used to fine-tune the LLM as the \our.

\subsection{Self-diversification for \our}

Dataset uniformity and diversity are essential to text classification \cite{text-bias} while the contents from LLMs are generally biased, especially when sampling from a single instruction \citep{llm-bias,bias-ai-content}. 
Thus, we propose a novel self-diversification technique to improve the generation quality from our \our. 
The main idea is to instruction-tune the LLM on the same instruction with several semantically different data samples. We refer to a pre-trained text embedder, specifically E5 \cite{E5}, to calculate the semantic similarity \cite{semantic-textual-similarity}. 
In our implementation, we reuse the instructions in the instruction-tuning dataset. 
For each instruction, we generate many (We select $1024$) training data\footnote{Generally, the data share the same label set.} and encode the data into the latent embedding space. 
As the data are formalized as Python dictionaries, we concatenate the embeddings of the values (texts) corresponding to a fixed order of keys.

$$E(d) = \bigoplus_{i=1}^{n} E(d[l_i])$$

\noindent where $E(\cdot)$, $d$, $l_i$ refer to the encoder, the data (dictionary) and the $i$-th label. $\oplus$ represents the concatenation operation and $n$ represents the total label number. 
After all data are encoded, we run a $K$-means (We select $K=8$) clustering algorithm on the embeddings and find the K samples with embeddings that are closest to the cluster centers. 
These samples, together with the instruction, establish a one-to-many mapping that maps instruction to very semantically diverse data samples. 
We incorporate these data in a batch of $K$ and further instruction-tune the LLM on it. 
Intuitively, this procedure will increase the appearance probability of data with unique semantics to benefit the incubated classifier. 

%% file: 4-exp.tex
\begin{table*}
\small
\centering
\begin{tabular}{lccccccccc}
\toprule
Method & \textbf{SST-2} & \textbf{SST-5} & \textbf{Emotion} & \textbf{AG News} & \textbf{NYT-LOC} & \textbf{TREC} & \textbf{SNIPS} & \textbf{Hillary} & \textbf{Average} \\
\midrule
Prompting& $91.43$ & $39.95$ & $46.65$ & $77.65$ & $71.07$ & $60.80$ & $42.29$ & $63.46$ & $61.66$ \\
\midrule
ZeroGen & $82.04$ & $39.37$ & $45.40$ & $65.57$ & $78.62$ & $59.10$ & $89.78$ & $57.97$ & $64.73$ \\
ProGen & $84.07$ & $41.49$ & $46.00$ & $67.72$ & $79.64$ & $59.80$ & $90.21$ & $57.42$ & $65.79$ \\
\our & $\textbf{90.01}$ & $\textbf{46.06}$ & $46.55$ & $69.46$ & $81.86$ & $\textbf{71.40}$ & $\textbf{93.57}$ & $\textbf{67.31}$ & $\textbf{70.78}$\\
$\quad$-Diversification & $85.45$ & $45.29$ & $\textbf{46.80}$ & $\textbf{69.91}$ & $\textbf{83.58}$ & $63.60$ & $91.07$ & $64.01$ & $68.71$ \\
\midrule
\our w/ GPT-4 & $86.99$ & $44.43$ & $47.80$ & $80.79$ & $86.87$ & $77.80$ & $94.14$ & $64.01$ & $72.85$ \\
\bottomrule
\end{tabular}
\caption{Text Classification Benchmark Results. All methods are based on LLaMA-2 except for \emph{\our w/ GPT-4}.} 
\label{tab:main}
\end{table*}

\section{Experiments}


We conduct several experiments to evaluate the performance of our \our. We include experiments on traditional datasets, and revised datasets with the label ``Other''. We also evaluate the ability of \our to handle complex personal labels and even ones with conjunctive relationships.

\subsection{Evaluations and Datasets}

Towards a comprehensive evaluation of our \our, we organize the evaluation into three scenarios.

\paragraph{(1) Traditional Benchmarks}
We include $8$ traditional text classification benchmarks, such as sentiment analysis classification (1) \textbf{SST-2}~\cite{sst}, (2) \textbf{SST-5}~\cite{sst}, and (3) \textbf{Emotion}~\cite{emotion}, topic classification (4) \textbf{AG News}~\cite{ag_news}, news location classification (5) \textbf{NYT-LOC}~\cite{nyt}, question type classification (6) \textbf{TREC}~\cite{trec}, intent classification (7) \textbf{SNIPS}~\cite{snips}, and (8) sentiment classification towards a particular public figure \textbf{Hillary}~\cite{tweeteval}.

\paragraph{(2) Label ``Other''}
We also test the ability of \our to handle stronger label interdependency by datasets with ``\emph{Other}''. 
We convert several datasets by grouping minor categories based on the proportion as a single ``Other'' label, with details mentioned in the Appendix~\ref{apdx:misc}.
These datasets include unbalanced datasets: Emotion, NYT-LOC, and \textbf{Massive}~\cite{massive}. 
These revised datasets will be also released for reproducibility

\paragraph{(3) Complicated Class Definitions}
To further showcase the usefulness of \our, we come with several complicated instructions for \our to incubate text classifiers that will be later used to mine the desired texts from massive raw documents, such as \textbf{TED Talk Summary}\footnote{\href{https://huggingface.co/datasets/gigant/ted_descriptions}{Huggingface: chirunder/gigant/ted\_descriptions}}, \textbf{Steam Game Description}\footnote{\href{https://huggingface.co/datasets/FronkonGames/steam-games-dataset}{Huggingface: FronkonGames/steam-games-dataset}}, and \textbf{Text Message}\footnote{\href{https://huggingface.co/datasets/chirunder/text_messages}{Huggingface: chirunder/text\_messages}}.

Note that all the datasets in our evaluations are \textbf{excluded} from the instruction-tuning data of \our.

\subsection{Implementation Details}

We implement \our by tuning LLaMA-2 (\texttt{LLaMA-2-7b-hf}) \cite{llama2} on our constructed instruction-tuning dataset with AdamW optimizer \cite{adamw} and cosine annealing learning rate scheduler~\cite{SGDR}. 
The specific hyperparameters for the optimization are shown in Table~\ref{tab:hyperparameter} in Appendix~\ref{apdx:hyperparameter}.

For all experiments, our \our is queried to generate $1024$ data samples to incubate a small classifier, which is selected as \texttt{RoBERTa-Large} \cite{roberta}. 
The RoBERTa is fine-tuned with the same optimizer and scheduler as for instruction-tuning and the hyperparameters for the incubation are also presented in Table~\ref{tab:hyperparameter}. 

\subsection{Compared Methods}

One can directly prompt the LLM, LLaMA-2 (\texttt{LLaMA-2-7b-hf}), which is the same as the LLM used in \our, with all the labels and the input text in the prompt and ask it to categorize the text into one of the labels~\cite{llm-tc}. 
We name this method as \textbf{Prompting}.

We include strong baselines that generate training data without requiring massive raw texts as follows. 
\begin{itemize}[nosep,leftmargin=*]
    \item \textbf{ZeroGen} \citep{zerogen}: This method prompts LLMs (\texttt{LLaMA-2-7b-hf}) to generate texts based on label descriptions. 
    Different from our \our, ZeroGen handles each label separately. 
    Towards a fair comparison with our method, we formalize our instruction-tuning dataset as the template used in ZeroGen to further fine-tune the model.
    \item \textbf{ProGen} \citep{progen}: This method further develops ZeroGen by an iterative ICL-based augmentation. 
    With the classifier obtained from ZeroGen, ProGen selects the most helpful data according to the decision boundary of the classifier. 
    These data are used as in-context examples to prompt the LLM to generate more helpful data to strengthen the classifier. 
\end{itemize}

\textbf{\our w/ GPT-4}: This is a variant of our \our that prompts \texttt{GPT-4} with in-context examples from the Huggingface platform and the instruction to sample the training data.
We present this not as a baseline but to showcase that the \our idea also applies to propriety LLMs. 

All data generation baselines generate the same amount of data ($1024$ per class) towards a fair comparison.

\subsection{Text Classification Benchmark Results}

The experiment results on traditional benchmarks are shown in Table~\ref{tab:main}. The comparison between ZeroGen and ProGen baselines verifies our \our has a significant advantage over those label interdependency-agnostic methods, which indicates the advantage of \our to consider the full label set in the instruction. 

Moreover, the self-diversification procedure is shown to highly contribute to the performance of \our, which boosts the performances on $5$ out of $8$ datasets and achieves comparable performances on others. Thus, self-diversification is verified to be a reliable and beneficial technique to strengthen the \our. 

We also present the performances of direct inference based on \texttt{LLaMA-2-7b-hf}, which is generally outperformed by the small LMs fine-tuned on datasets generated by LLMs. This result is consistent with the discovery that LLMs are better generators than discriminators \citep{auggpt}. This further supports incubating a small model for text classification from the LLM rather than directly prompting the LLM, not only for efficiency but also for performance improvement. 

Finally, we evaluate the ICL-based \our with GPT-4 as the backbone model. With a significantly larger amount of parameters, \our with GPT-4 outperforms the one based on LLaMA-2. 
This indicates larger backbone models can further scale up the performance of our \our. 
Also, tunable models can benefit from self-diversification to narrow the gap between the close-source GPT-4, which can also be improved once it becomes open-source for fine-tuning.

\subsection{Label ``Other'' Results}

\begin{table}
\small
\centering
\scalebox{0.975}{\begin{tabular}{lcccccccc}
\toprule
Method & \textbf{Emotion} & \textbf{NYT-LOC} & \textbf{Massive} \\
\midrule
Prompting & $43.15$ & $62.11$ & $57.67$ \\
\midrule
ZeroGen & $52.65$ & $69.27$ & $56.46$ \\
ProGen & $52.80$ & $69.64$ & $57.16$ \\
\our & $\textbf{56.00}$ & $\textbf{84.19}$ & $\textbf{68.36}$ \\
$\quad$- Diversification & $55.00$ & $76.39$ & $61.53$ \\
\midrule
\our w/ GPT-4  & $53.40$ & $78.36$ & $73.84$ \\
\bottomrule
\end{tabular}}
\caption{Results on datasets with the ``Other'' class. } 
\label{tab:misc}
\end{table}

We present the experiment results on datasets with miscellaneous (label ``Other'') in Table~\ref{tab:misc}. 
The awareness of the miscellaneous category is important for classification \citep{misc-classification}, especially when limited labels are known in a large corpus.
For ZeroGen or ProGen, we use the label name ``\emph{Other than ... (other labels)}'' to prompt for generation. 
We can observe a significantly larger gap between the \our and the label interdependency-agnostic methods, which shows the advantage of \our on datasets with miscellaneous. 
Furthermore, the self-diversification shows a more prominent improvement in performance. 
This phenomenon can be attributed to the requirement for a more diverse generation by the miscellaneous category.

\begin{table*}[t]
\small
\centering
\begin{tabular}{lclclclc}
\toprule
Target & TED Summary & Target & Steam Game & Target & Text Message \\
\midrule
``About AI'' & $100\%/100\%$ & ``Action'' & $90\%/90\%$ & ``Positive'' & $98\%/98\%$ \\
``About Climate'' & $100\%/100\%$ & ``RTS'' & $74\%/77\%$ & ``Request'' & $97\%/98\%$ \\
``By Educator'' & $94\%/94\%$ & ``Card'' & $100\%$/$100\%$ & ``About Food'' & $98\%/98\%$ \\
``Funny'' & $75\%/80\%$ & ``Relaxing'' & $100\%/100\%$ & ``Work-related'' & $83\%/86\%$ \\
\bottomrule
\end{tabular}
\caption{Precision@$100$ (GPT-4 Evaluation/Human Evaluation) of incubated retrievers on unannotated corpora.} 
\label{tab:mine}
\end{table*}

\begin{table*}[t]
\small
\centering
\begin{tabular}{llcc}
\toprule
Logic &  Target & Direct Incubation & Conjuctive Incubation\\
\midrule
$L_1 \land L_2$ & ``Positive and about food'' & $85\%/85\%$ & $\textbf{88\%}/\textbf{88\%}$ \\
$L_1 \lor L_2$ & ``Positive or negative'' & $99\%/99\%$ & $\textbf{100\%}/\textbf{100\%}$\ \\
$L_1 \land \lnot L_2$ & ``Positive but not excited'' & $74\%/72\%$ & $\textbf{89\%}/\textbf{86\%}$ \\
$L_1 \land L_2 \land L_3$ & ``Positive, about food, and with dish name'' & $40\%/43\%$ & $\textbf{84\%}/\textbf{85\%}$ \\
\bottomrule
\end{tabular}
\caption{The performance of incubated retrievers with logical conjunctions.} 
\label{tab:logic_mine}
\end{table*}

\subsection{Complicated Class Definition Results}

We further showcase how \our can be applied to satisfy personal demands, such as mining items preferred by an individual. For each raw corpus, we propose four attributes a user might be interested in, such as ``About AI'' for TED Talks. For each attribute, we create an instruction to build a text classifier with two labels: the target attribute and the miscellaneous label ``Other''. 
We use the incubated classifier to score each raw text and select the texts with the top scores. 
For evaluation, we ask GPT-4 and humans whether the mined texts satisfy the demand with Precision@100 as the metric.

The text mining performance is presented in Table~\ref{tab:mine}. 
\our incubates strong text miners with generally high precision on all setups. Remarkably, we achieve nearly or exactly $100\%$ precision on several targets. Moreover, our miners are validated to be able to handle different text domains, enabling a broad application of our \our. 

\subsection{Incubation with Logical Conjunction}


We further showcase how to utilize \our to satisfy more complicated user demands. We increase the label complexity by adding logical conjunctions into labels, that are ``and'' ($\land$),  ``or'' ($\lor$), and ``not'' ($\lnot$). The logical conjunctions represent a finer-grained demand from the user. For instance, one may want to search for texts that are ``Positive and about food'', as ``Positive'' $\land$ ``About food''. 

To realize such finer-grained text mining, we utilize the maneuverability of \our to incubate multiple text miners and combine their scores with logical probabilistic calculations as follows,

\begin{itemize}[nosep,leftmargin=*]
\small
    \item $P(L_A \land L_B) = P(L_A) P(L_B)$
    \item $P(L_A \lor L_B) = P(L_A) + P(L_B) - P(L_A \land L_B)$
    \item $P(\lnot L_A) = 1 - P(L_A)$
\end{itemize}

\noindent where $L_A$, $L_B$ are two labels used as the targets for the incubation. Here we view the labels as independent for simplification. We use the Text Message corpus for text mining. For evaluation, we keep the previous scenario unchanged. We compare two types of incubation scenarios,

\begin{itemize}[nosep,leftmargin=*]
    \item \textbf{Direct Incubation} \our only incubates one text miner with the full label name, such as ``Positive and about food''.
    \item \textbf{Conjunctive Incubation} first decomposes the label name into multiple ones with corresponding conjunctions, like decomposing ``Positive and about food'' into ``Positive'' $\land$ ``About food''. Then the score is calculated based on logical probabilistic calculations. 
\end{itemize}

The experiment results are presented in Table~\ref{tab:logic_mine}. Conjunctive incubation generally outperforms direct incubation, which shows the benefit of this strategy. As conjunctive incubation also shows strong capability on three logical variables, this shows how \our can be customized to more complex settings.

\begin{table*}[t]
\small
\centering
\begin{tabular}{llll}
\toprule
 Target & Generated data with target label & Generated data with misc label\\
\midrule
& Hey! I love the new update. It's awesome! & Just checking in on the progress of the project. \\
``Positive'' & Wow, you got the tickets for our dream holiday! & I've booked the flights for next week. \\
 & I absolutely love the new design of the app. & I'm having trouble logging into my account. \\
 \midrule
 
& Can you send me the report by end of today? & What did you do during the weekend? \\
``Request'' & Could you please bring me a coffee? & How was your day? \\
 & Can you pass me the salt? & Hey, did you catch the game last night? \\

 \midrule
 
& The pizza at Mario’s is the best in town! & I have an important meeting at 10am tomorrow. \\
``About food'' & I'm craving for a burger and fries! & I might go for a run later. \\
 & I just tried that new sushi place. Totally worth it! & Hey, what time does the movie start? \\

 \midrule
 
& We need to finalize the report by tomorrow. & Hey, do you want to catch a movie tonight? \\
``Work-related'' & The meeting is scheduled at 3 PM tomorrow. & Do you want to catch up for dinner tonight? \\
 & The project deadline has been extended. & Hey! What are you up to this weekend? \\
 
\bottomrule
\end{tabular}
\caption{The performance of incubated retrievers with logical conjunctions.} 
\label{tab:generate}
\end{table*}

\subsection{Case Studies on Generated Training Data}

To more concretely demonstrate the intermediate processes in the incubation, we launch a study on the generated texts from the \our for classifier incubation. We demonstrate the generated training data for data mining in the text message corpus in Table~\ref{tab:generate}. For each column, there is a piece of text generated with the target value and the other one in the same Python dictionary with the miscellaneous label ``Other''. 

The most straightforward observation is the generated data correctly follows the label, which guarantees the foundational precision of the incubated classifiers. Also, the generated texts incorporate a wide range of syntactic structures and semantic contents for the training data diversity. For the miscellaneous label, we can observe the \our to cover various potential negative labels. For instance, the miscellaneous category for ``About food'' includes labels such as ``About meeting'', ``About sports'', ``About movie'', which broadens the negative set understood by the incubated classifier.

Finally, we can view some attribute correlations between the data in the same generated Python dictionary. In the ``Positive'' example, the three samples have the same topic ``Project'', ``Travel'', and ``App''. With these data different in the target attribute but same in other attributes, the incubated classifier can better focus on the target attribute and eliminate spurious correlations.

%% file: 5-analysis.tex
\section{Analyses of \our}

We analyze the properties of the \our to deepen the understanding and guide the usage.

\subsection{Incubation Dataset Size}

\begin{figure}
    \centering
    \includegraphics[width=\linewidth]{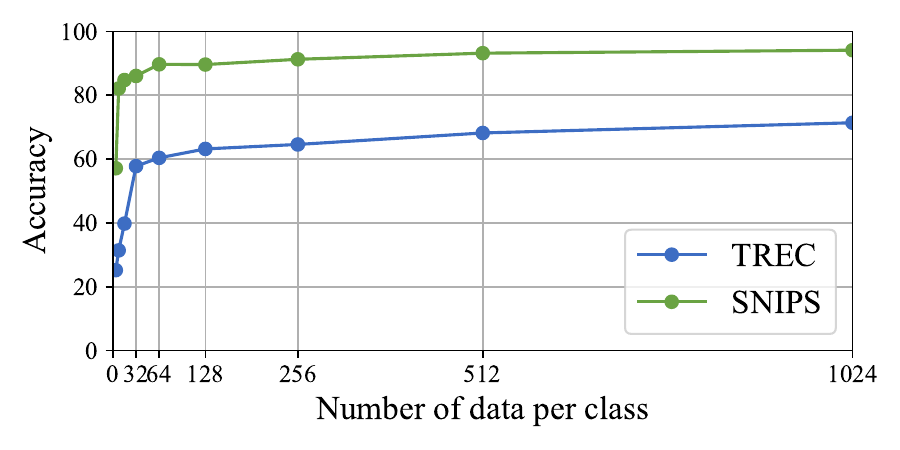}
    \vspace{-10mm}
    \caption{Incubation dataset size analysis.}
    \label{fig:size}
\end{figure}

We first adjust the number of data generated from \our to investigate how the incubated classifier will be affected. We conduct experiments on TREC and SNIPS datasets with incubation data size from $4$ to $1024$. The results are illustrated in Figure~\ref{fig:size}. From the shown scaling-up trend, there is a clear threshold ($64$) on the dataset size, after which the performance gained from generating more training data becomes limited. Thus, we recommend \our users generate at least $64$ data samples for the classifier incubation.

\subsection{Instruction Robustness}

\begin{figure}
    \centering
    \includegraphics[width=\linewidth]{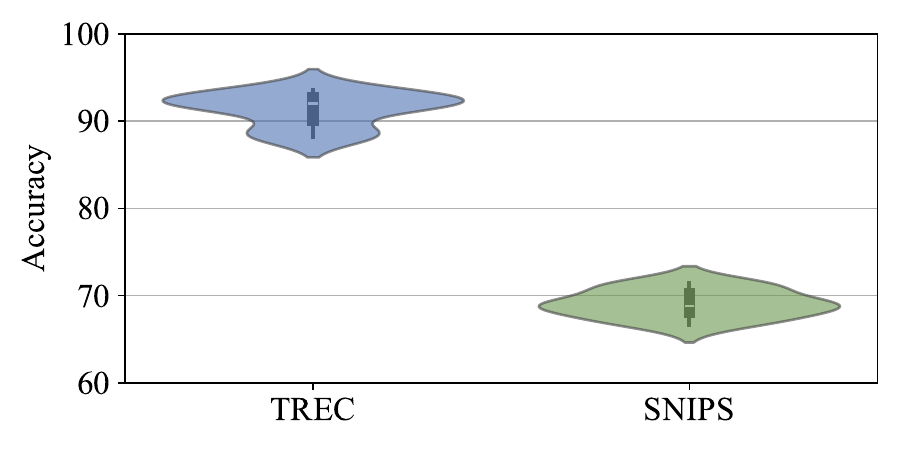}
    \vspace{-10mm}
    \caption{Analysis of \our instruction robustness.}
    \label{fig:robust}
\end{figure}

We then check the robustness of \our to instructions by testing with different but semantically equal instructions. We rephrase each instruction for TREC and SNIPS into $10$ different versions and then run the incubation pipeline for evaluation. 

The robustness evaluation is presented in Figure~\ref{fig:robust}. We can observe the lexical and syntactical attributes, which is changed in the rephrasing, have limited impact on the incubated result. Thus, we conclude our \our is robust against the variations of the same instruction. 

\subsection{Efficiency Analysis}

\begin{figure}
    \centering
    \includegraphics[width=\linewidth]{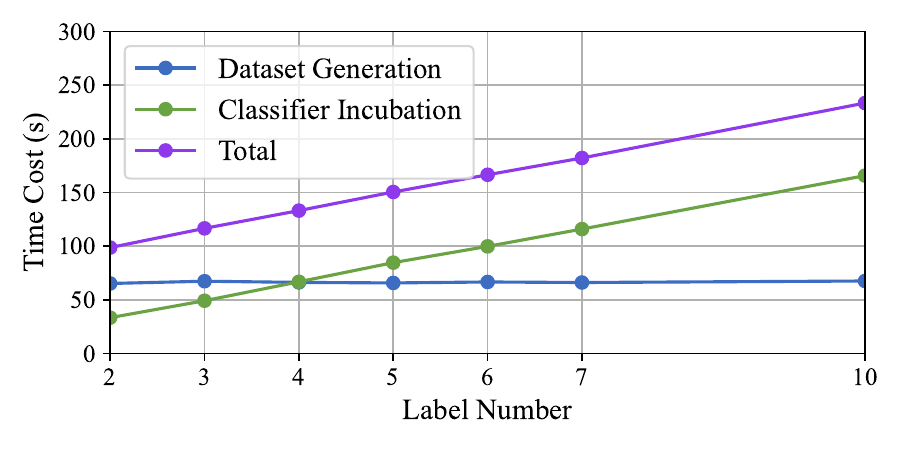}
    \vspace{-10mm}
    \caption{Efficiency Analysis of \our instruction.}
    \vspace{-5mm}
    \label{fig:time}
\end{figure}

We analyze the time efficiency of the \our to explore its efficiency in deployment. For dataset generation, we run the LLaMA model with the acceleration by the \texttt{vllm} package \citep{vllm}. For the small classifier incubation, we fine-tune the model with the trainer in the \texttt{transformers} package \citep{transformers}. We evaluate the time for dataset generation and classifier incubation (fine-tuning). The time is obtained by averaging the results in experiments on the $8$ traditional benchmarks, which is illustrated in Figure~\ref{fig:time}. All experiments are run on a single A100 device. 

For dataset generation, the average time is $67.53$s. The generation times for all benchmarks are distributed around this average since \texttt{vllm} has a fixed max length limitation for decoding. For classifier incubation, the time is almost linearly dependent on the number of labels, which shows an average of $15.16$s time cost per class. 

Thus, the time efficiency of our \our is feasible to incubate personal classifiers. Also, the main time cost happens in classifier incubation rather than calling the LLM for dataset generation, especially when the label number is large.

%% file: 6-con.tex
\section{Conclusion and Future Work}

In summary, this paper proposes a new framework for model incubation by querying an instruction-tuned LLM. Our model, \our, is pre-trained on Huggingface resources and ICL-based augmentation. The \our is further strengthened by a novel self-diversification technique with the help of text embedders. We show the \our to be able to incubate strong classifiers for traditional benchmarks and customized text mining, following the demand written in the input instructions. We also include comprehensive analysis to explore the properties of the \our for deeper understanding and better application guidance.

Future work will concentrate on two tracks. 1) \textbf{Improve the incubation quality:} We can incorporate existing or new methods to improve data generation quality like higher diversity and harder negative samples. 2) \textbf{Broaden the scope of incubated models:} The incubated model can be more than classifiers, such as question responder and text summarizer. These models might require more complicated instruction understanding and other techniques for model enhancement. 

%% file: 7-lim.tex
\section*{Limitation}

While \our shows strong performance in producing reliable and customized classifiers, it has some limitations that can be further improved in future works.

\paragraph{Instruction Effort:} Current \our requires the user to include all label names in the instruction, which adds effort for the user to create instructions, especially when the label number is large or the user is unclear about the label names. A combination with existing work \cite{wot-class} might be a direction to reduce user efforts further. 

\paragraph{LLM Knowledge Dependence:} As an LLM-only methods, the \our is only able to generate text within its knowledge scope. For emerging labels, the \our still has to rely on delicate explanations or in-context examples to handle them.